\documentclass[11pt,letterpaper]{article}
\usepackage{cogsys}
\usepackage[T1]{fontenc}
\usepackage{times}
\usepackage[pdftex]{graphicx} %
\usepackage{wrapfig}
\usepackage{caption}

\usepackage{natbib}
\begin{document} 

\title{Improving Language Model Prompting \\ in Support of Semi-autonomous Task Learning}
 
\author{James R. Kirk}{james.kirk@cic.iqmri.org}
\author{Robert E. Wray}{robert.wray@cic.iqmri.org}
\author{Peter Lindes}{peter.lindes@cic.iqmri.org}
\author{John E. Laird}{john.laird@cic.iqmri.org}
\address{The Center for Integrated Cognition, IQM Research Institute, 
         Ann Arbor, MI 48106 USA}
\vskip 0.2in
 
\begin{abstract}Large language models (LLMs) offer a potential source of knowledge for agents that need to acquire new task competencies within a performance environment. We describe efforts toward a novel agent capability that can construct cues (or ``prompts'') that result in useful LLM responses for an
agent learning a new task. Importantly, responses must not only be ``reasonable'' (a measure used commonly in research on knowledge extraction from LLMs) but also must be specific to the agent's task context and in a form that the agent can interpret given its native language capacities. We summarize a series of empirical investigations of agent prompting strategies and evaluate LLM responses against the goals of targeted and actionable responses for task learning. Our results demonstrate that actionable task knowledge can be obtained from LLMs in support of online agent task learning.
\end{abstract}

\section{Introduction}

Large language models \citep[LLMs;][]{brown_language_2020,bommasani_opportunities_2021_local} offer a potential source of knowledge for autonomous agents, especially those that learn new tasks within a performance environment \citep{wray_language_2021_local}. Consider a robotic agent in an office environment. It would likely have built-in capabilities, such as delivering and fetching, but also a capacity to learn new tasks specific to its particular office environment. Examples might include tidying a kitchen or conference room, monitoring building security, and performing simple maintenance tasks. While it is feasible today for agents to learn tasks such as these via interactive task learning \citep[ITL;][]{gluck_interactive_2019_special}, the required instruction can be tedious for humans, requiring time and concentration to teach a new task. Additional sources of knowledge, such as LLMs, offer potential to speed agent knowledge acquisition, reducing the time and effort needed to instruct a new task.

The class of LLMs we explore, \emph{decoders}, are sentence-completion engines. Decoder models generate output text in response to a text input. For illustration, in Table~\ref{tab:prompt_response_introduction}, two examples of inputs (or ``prompts'' in the vernacular of LLMs) are shown. Both of these prompts are natural language sentences that prompt the LLM about tidying a conference room. These sentences are input to the LLM in plain text. We use the GPT-3 LLM \citep{brown_language_2020} here and throughout the paper. GPT-3 produces responses to these inputs, also in plain text. A ``temperature'' parameter can be manipulated that encourages higher or lower likelihood choices of the next token (temperature=0 means that the highest likelihood option is always chosen). As suggested by these responses, GPT-3 produces responses that (to a human reader) provides relevant information tidying a conference room. Our primary research goal is to harness such latent knowledge within an LLM so that an agent can acquire new knowledge from LLMs to support autonomous task learning.

This goal is not unique. Knowledge extraction from LLMs has emerged as a sub-field within the LLM community, demonstrating that physical \citep{bosselut_comet_2019_local}, social \citep{bosselut_dynamic_2021_local}, and even narrative \citep{mostafazadeh_glucose_2020_local} knowledge can be extracted from LLMs. As seen from the examples in Table~\ref{tab:prompt_response_introduction}, seemingly small changes in the prompt can lead to significantly different responses.  Thus, a key challenge of knowledge extraction from LLMs is determining how to form an appropriate prompt so that the resulting response is useful \citep[``prompt programming,''][]{reynolds_prompt_2021_local}. 

Knowledge extraction via LLMs has generally focused on the acquisition of general world knowledge without anticipating direct application of the knowledge to a specific situation. For example, COMET \citep{bosselut_comet_2019_local} was trained to produce assertions compatible with ConceptNet \citep{speer_conceptnet_2017}, a general knowledge base. New assertions from COMET could extend ConceptNet's coverage of world knowledge, but the extraction process included no consideration of how that extracted knowledge might be used in a specific circumstance by an agent in the future. However, an embodied robot requires responses specific to its situation and its affordances. Further, robot affordances will often differ from the (typically human) embodiment that is (usually) implicit in the LLM training corpus \citep{ahn_as_2022_local}. Thus, a new requirement for knowledge extraction for autonomous agent learning is construction of prompts  that deliver LLM responses with sufficient specificity to the agent's current situation.

\begin{table*}
    \centering
    \begin{tabular}{p{0.47\textwidth} p{0.47\textwidth}}
    \hline 
    \textbf{Prompt} (Plain-text input to GPT-3) & \textbf{Response} (Plain-text output from GPT-3)\\
    \hline
    How do you tidy a conference room? 
     & To tidy a conference room, you will need to gather all of the trash and dispose of it, wipe down all of the surfaces, and vacuum the floor. \\
    \hline 
    What tasks would you do to tidy a  & -Wipe down all surfaces \\
    conference room? & -Vacuum the floor \\
             & -Dust \\
             & -Empty the trash \\
             & -Wash the windows \\
             & -Restock supplies \\
\hline
    \end{tabular}
    \caption{Examples of prompts (input) and responses (output) from GPT-3 (Temperature=0, model=text-davinci-002).}
    \label{tab:prompt_response_introduction}
\end{table*}

For an agent to use LLM responses, it must also be able to parse and to ground them to its current situation using the native language understanding facilities it has. Referring again to Table~\ref{tab:prompt_response_introduction}, some phrases/steps are straightforward to interpret (``empty the trash'') but others might be much more challenging for an agent to ground and to act on (e.g., ``gather all of the trash and dispose of it''). Further, the second response, which is generated by the LLM as a bulleted list, can be parsed and processed individually in comparison to the more complex sentence produced in response to the first prompt. 

A response that cannot be interpreted with the agent's native language facilities, even if it is appropriate and interpretable for a human reader, does not contribute to the agent's attempt to learn a new task. Thus, another requirement for an embodied agent to be able to use the LLM for knowledge acquisition is that the agent produce prompts (inputs) to the LLM that result in responses that the agent can parse and ground (interpret).

To summarize, LLMs offer potential as a knowledge source, but to exploit them, an agent will need to construct effective prompts. In this paper, we summarize a systematic exploration of prompt construction strategies, taking into account  specific requirements of an embodied agent as the producer of prompts to the LLM and as the consumer of the subsequent responses. We describe variations in the construction and parameterization of these prompts and evaluate the responses given those experimental conditions. We also discuss directions toward further improving agent-based prompt construction, learning task goals, and complementary methods to verify responses from LLMs. Long-term, the prompt strategies presented in this paper demonstrate that future agents can construct effective, customized prompts for their specific task-learning needs and thus extract actionable knowledge from LLMs.

\section{Context: Online Task Learning}
As suggested above, online, situated agent learning presents specific challenges and requirements for the responses from an LLM, which, in turn, create requirements for the prompts produced by an agent. In this section, we summarize prior work toward an online, task-learning agent that learns from human interaction, introduce a specific example of task learning, and then describe measures for assessing the performance of prompt engineering methods.

\subsection{Online Task Learning via Human Interaction}
This research builds on a prior ITL agent, developed in the Soar cognitive architecture, that learns novel tasks via natural language interaction with a human instructor \citep{kirk_learning_2019,mininger_expanding_2021}. The agent and instructor participate in a dialog in which the agent asks to be taught a new task, asks for goals and subtasks associated with the new task, and seeks to connect or ``ground'' each instruction to its current understanding of the situation. The agent has natural language parsing, interpretation, and generation capabilities that allow it to participate in a restricted but still expressive dialogue of interactions. In particular, the agent seeks clarification whenever it cannot ground a phrase or term in the instruction.

The agent uses iterative-deepening planning to evaluate if it can satisfy task goals without further instruction. For example, if it has a goal that the ``bottle is in the recycling bin,'' planning finds a simple, two-step plan: \texttt{pick-up(bottle)}, \texttt{put-down(bottle,recycling-bin)}. Further, the agent uses subtask decomposition and planning to determine how to accomplish effector-level task steps. Thus, \texttt{pick-up(bottle)} is achieved by first executing \texttt{approach(bottle)} and \texttt{face(bottle),} eliminating the need for low-level instructions.

These capabilities and the resulting process lead to a step-by-step learning of the task. The learning occurs situated within an environment (online) and a successful interaction leads to the agent learning to perform the task after a single learning interaction (``one-shot''). 

This agent provides a foundation and a context for the research that follows. Rather than attempting to learn solely from an instructor, this agent will instead employ the LLM as a proxy for the instructor (e.g., ask the LLM for the steps in a task rather than a human instructor) while also using its planning knowledge. This design commitment has two important ramifications for this work. First, although a human instructor has knowledge and context that they can draw on to respond to agent requests, the LLM (generally) lacks this knowledge. Therefore, the agent must provide sufficient context in its prompts to achieve actionable responses. Second, the native language capabilities of the agent are limited (which human instructors readily adapt to). Thus, in addition to providing context, the agent must also bias generation of responses toward ones that the agent can actually parse and ground.

In summary, this approach offers the obvious benefit of building on the existing agent capabilities, but it also imposes additional requirements. We discuss the specific requirements for task-learning from an LLM after first outlining a specific learning task.

\subsection{Example Task: Tidying a Conference Room}

In this section, we introduce a specific learning task, which we use both for illustration and later for experimentation. Imagine a robot designed to support staff and operations in a generic office building. Such a robot would likely have built in functions, such as delivering or fetching mail, packages, and other small items (e.g., lunches, photocopies) for workers in the building. The robot would need not only to learn the specific configuration of the building (who is located in which office), it would also need to learn new tasks specific to this office setting. 

\begin{table*}[b]
    \centering
    \begin{tabular}{p{0.15\textwidth} p{0.7\textwidth}}
    \hline 
    Object & Object Properties \\ \hline \hline
    can & contents empty; material metal; property soda; in conference room \\
bottle & contents full; material plastic; in conference room \\
cup & contents empty; material paper; in conference room \\
cup & material glass; in conference room \\
mug & material ceramic; property coffee; in conference room \\
table & in conference room \\
chair & in conference room \\
recycling bin & in conference room \\
waste bin & in conference room \\
\hline
    \end{tabular}
    \caption{Objects and features used for the ``tidy conference room'' task.}
    \label{tab:conference_room_objects}
\end{table*}

We focus on one such task, learning to tidy up a conference room within the building. The task is specific to the building in a few ways, including the desired configuration of the conference room and specific expectations (rules and norms) about that configuration (e.g., should whiteboards be erased?). We focus on learning those aspects of the task that are not so tightly bound to the specific conference room, yet that may have some location- and context-specific conditions. To illustrate, consider the list of items enumerated in Table~\ref{tab:conference_room_objects}. The table lists the name of the object as well as the properties of the object provided in the agent representation of this environment. What should the robot do when is encounters bottles, cups, etc. in the conference room when tidying? We can imagine that a full plastic bottle should be emptied and recycled, provided there is a place to empty the bottle and recycling is available. A paper cup should likely be put in a waste bin, but a glass cup should be cleaned, and so on. We will use this task throughout the rest of the paper to illustrate concepts and as a task for actual experiments.

We assume learning task steps such as these from the LLM would allow the agent to gain some basic ``tidying'' task knowledge. For example, if the agent can learn to recycle bottles it finds in the conference room, or even to propose recycling the bottle to the instructor as a potential action, the instructor's task in teaching the robot to tidy a specific conference room will be simpler.  In other words, we envision the role of the LLM to be analogous to the role of planning outlined above; the knowledge obtained from the LLM will reduce the role of the human in the interaction, making it less tedious and much faster, as planning allows the human to skip the specification of tedious steps like ``approach the can.'' It may be possible in the longer term to learn new tasks from the LLM without any human interaction by using additional knowledge sources. However, the larger goal of our effort is ``semi-autonomous'' learning from the LLM where the instructor can confirm or disconfirm LLM recommendations rather than requiring fully correct and situationally-appropriate responses.

\subsection{Task-learning Requirements for LLM Responses}

Online task learning imposes requirements on the results (responses) received from the LLM. We focus on three: reasonableness, situational relevance, and interpretability. These requirements will form the basis of our evaluation of alternative prompting strategies.

\subsubsection{Reasonableness}
A reasonable response is one that humans would generally deem acceptable as an instruction in the task environment. For \texttt{tidy-conference-room}, suggestions such as ``clearing the table,'' ``erasing the whiteboards,'' or ``returning the coffee pot to the kitchen'' would be reasonable for tidying conference rooms we imagine. However, a suggestion such as ``return the crib to the bedroom,'' or ``remove the outlet cover'' are unlikely or atypical steps one would encounter when tidying a conference room. Cribs are not typically encountered in conference rooms (or an office setting more generally), and outlets are often present in conference rooms, but removing an outlet cover is a maintenance or repair step rather than part of tidying. 

The challenge for the agent is to construct prompts that bias the resulting responses toward reasonableness. As suggested by the counterexamples, the prompt must reference the context (such as a conference room) and the task (tidying) so that the LLM's responses are consistent with both.

Reasonableness  is sometimes called ``accuracy'' in 
work on knowledge extraction \citep{bosselut_comet_2019_local}, but, because a robot requires even more specificity in LLM responses (next section), 
we prefer the term ``reasonableness'' over ``accuracy''  to avoid labeling responses as ``accurate'' when they may not be actionable for the robot in its specific situation.

\subsubsection{Situational Relevance} 
Reasonableness alone is not sufficient to ensure utility for task learning; the LLM response must be relevant to the agent's specific situation. This requirement is potentially more challenging than the previous one because, while the statistical patterns of text generation from an LLM can produce reasonable responses, only a small fraction of those responses are likely to be directly relevant to the specific circumstances of the agent. 
We identify two distinct components of this requirement:

\emph{Matched Embodiment:} An agent's embodiment in its environment and the affordances provided by that environment may differ from the human embodiment implicit in the majority of text data used to train an LLM. Our (simulated) robot has a single arm and gripper, so that responses that are easily executed by most humans (``open the fridge door while holding a can'') are not possible for the robot. Learning to generate prompts that match a robot's embodiment has recently emerged as a research direction for using LLMs with robots \citep{ahn_as_2022_local,logeswaran_few-shot_2022_local}.

\emph{Situational Specificity:} Above we described ``erasing the whiteboards'' as a reasonable step to take when tidying a conference room. However, not all conference rooms have whiteboards. Further, in some office contexts, erasing a whiteboard may be strongly discouraged. While it might be feasible to incorporate these constraints via fine-tuning for the task and domain, in general the LLM lacks inherent knowledge of the agent's specific situation. Fine-tuning also requires foreknowledge of the agent's situations, which  can be difficult to predict, especially when learning in novel situations, which is the objective of learning new tasks online. Thus, the agent must construct prompts that engender responses to the agent's specific task context.

\subsubsection{Interpretability}

LLMs can generate open-ended, complex natural language utterances that exceed the ability of the agent's parsing and comprehension capabilities. Further, the LLM can introduce terms and concepts that are not known to the agent. Thus, another requirement is that LLM responses are interpretable by the agent. %

This requirement comes directly from an integrated, cognitive systems perspective on task learning. Today, any agent is limited to processing some restricted natural language (i.e., whatever its language capacities are, those ``native language capabilties'' are not as developed as typical adult speakers and users of that language). 
As we shall see below, an agent can bias generation toward the agent's native language capabilities through careful prompt construction.  This requirement is thus analogous to the way in which humans adapt their own language when encountering other actors with more limited language competencies (e.g., artificial agents, young children).

\section{Prompt Engineering}
Given a new task to learn, a specific context, and a particular embodiment, the agent's goal is to construct a prompt that results in a response that achieves acceptable reasonableness, relevance, and interpretability. The exact criterion that is sufficient for achieving ``acceptability'' will be specific to a given application. In this paper, we identify prompt engineering strategies that progressively improve performance on the measures rather than setting minimally acceptable values for them.

The strategies we discuss extend an earlier template-based prompting strategy \citep{wray_language_2021_local}. In that approach, general templates guide prompt construction by identifying various types of information that the agent needs to provide to the LLM. Table~\ref{tab:prompt_template} illustrates the basic template we use.\footnote{The ``full context'' variation includes additional delimiters and is located separately from the task definition. See section \ref{section:context} for a description.} Items in \textit{italic} represent plain text delimiters in the prompt text that we have introduced to help differentiate components of the prompt. The delimiters have no special meaning or significance to the LLM. Items in \textbf{bold} are further decomposed in later lines.

\begin{table*}
    \centering
    \begin{tabular}{p{.2\textwidth} p{.75\textwidth}}
    \hline 
    Component & 
    Specification/Description \\ \hline \hline
    \textbf{PROMPT} & \textit{(EXAMPLE)} \textbf{Example} \textit{(END EXAMPLES) (TASK)} \textbf{Partial-Task} \\
\textbf{EXAMPLE} & \textit{(TASK)} \textbf{Task} \textit{(END TASK)} \\
\textbf{TASK} & \textit{Goal:} \textbf{Goal}$_{style}$ \textit{Task context:} \textbf{Context}$_{style, context}$ \textit{Steps:}  \textbf{Steps}$_{style}$ \\
\textbf{GOAL} & Description of the task goal in the prescribed style \\
\textbf{CONTEXT} & Description of the situation in the prescribed style and context variations. See Table~\ref{tab:lang_styles} for examples of context templates in different styles. \\
\textbf{STEPS} & Enumerated list of task steps in the prescribed style \\
\textbf{PARTIAL-TASK} & \textit{Goal:} \textit{Goal}$_{style}$ \textit{Task context:} \textit{Context}$_{style,context}$ \textit{Steps:} 1. \\ 
\hline
    \end{tabular}
    \caption{The basic prompt template used in the experiments.}
    \label{tab:prompt_template}
\end{table*}

A \textbf{PROMPT} thus consists of a number of examples, followed by a partial task description. The partial task description includes a task goal description and a context description in a particular style, and the plain text delimiter \textit{Steps:} followed by a ``1.'' Because decoder LLMs are completion engines, this prompt template, when instantiated, drives the LLM to complete the partial task description, enumerating the steps needed to perform the task. Within an \textbf{EXAMPLE}, the LLM is given a full description of one or more example tasks, including the steps needed to execute each task.

Table~\ref{tab:lang_styles}, discussed in the next section on style variations, illustrates how the agent can instantiate the partial-task template with different objects (?object) observed in the environment and information related to the object, including its features (?feature-list) and location (?object-location). The templates are not specific to the task to be learned or to the specific domain, but they do embed some assumptions about task learning in general, such as attending to individual objects. 

One of the main roles of the templates is to enable the agent to embed specific environmental features in its prompts and thus facilitate greater situational relevancy. However, we have not yet performed systematic exploration of different templates, although we have explored some variations. Instead, we have developed this basic, general template and explored how the agent can customize and extend prompts from the foundation provided by this template.

\begin{table*}
    \centering
    \begin{tabular}{p{0.2\textwidth} p{0.78\textwidth}}
    \hline 
    Name & Variation \\ \hline \hline
    Style of Language & \textbf{Terse}, colloquial, predicate \\
    Delimiters &  Keywords (Paired parentheses, Unitary colon) \\ 
    Examples & Number = \{0, \textbf{1}, \textbf{2}, 3\} \\
    Situational Context & None, \textbf{Partial Context}, Full context \\
    Object Context & Name only, Name and properties \\    
    Stochasticity & Temperature = \{0, 0.3, 0.8\}
\\
\hline
    \end{tabular}
    \caption{Prompt engineering strategies explored in this analysis and exploration.}
    \label{tab:prompting_variations}
\end{table*}

Table~\ref{tab:prompting_variations} lists the different prompting strategies we explore to achieve the overall goal using template-based prompting as the foundation. Each row in the table names the specific method or component and enumerates variations of that method we report in this paper. Where we determined that a specific variation was always preferable, that variation is listed in \textbf{bold} type. In the remainder of this section, we discuss these strategies and variations in further detail.

\subsection{Language style}

To achieve more interpretable responses, we explored many variations in language styles for the prompts, leveraging the ability of GPT-3 to mimic the style or ``voice'' of a prompt when generating a response to it. Table~\ref{tab:lang_styles} displays prompt templates for three contrasting language styles for the example task of ``tidy conference room.'' The colloquial language style attempts to match how a person might  ask for task steps. The terse style uses a simple language with keyword delimiters, such as  ``Goal:'' and ``Steps:''.\footnote{As above, ``Steps:" is used at the end of a prompt to elicit responses from the LLM as a series of steps.} The predicate  style specifies the task and object relations as predicates with 1 or 2 arguments, such as ``Located-on(can, table).'' We included the predicate style based on some limited but promising success by others using a predicate formulation with GPT-3 for agent planning \citep{olmo_gpt3--plan_2021_local}.

\begin{table*}
    \centering
    \begin{tabular}{p{0.1\textwidth} p{0.85\textwidth}}
    \hline 
    \emph{Colloquial}     &   I see an \texttt{?feature-list} \texttt{?object} in \texttt{?named-location}. What are steps to tidy conference room with \texttt{?object} in it? 
\\[0.25ex] \cline{2-2} \emph{Terse}      &  Goal: tidy conference room. Task context: I am in \texttt{?named-location}. Aware of \texttt{?object}, \texttt{?feature-list}, \texttt{?object-location}. Steps: 
\\[0.25ex] \cline{2-2} \emph{Predicate}    &  Tidy(conference room). Observe(\texttt{?object}). Located-on(\texttt{?object}, \texttt{?object-location}). \texttt{?feature-list}. Steps: \\
\hline
    \end{tabular}
    \caption{Context templates for tidy conference room task with different language styles. (``?'' indicates the slots that are filled in by the agent during prompt construction.) These templates are used within the TASK and PARTIAL-TASK expressions as defined in Table~\ref{tab:prompt_template}.}
    \label{tab:lang_styles}
\end{table*}

We evaluated these variations for the conference room task. The terse language style influences GPT-3 to generate steps using simpler, more direct language that is more immediately interpretable by an agent with basic NLP capabilities, without any conversion or translation. Table~\ref{tab:gpt_Example} shows a prompt instantiated from the template in Table~\ref{tab:prompt_template} for a metal soda can using the terse format. As with all future examples, the prompt presented to GPT-3 is displayed above the horizontal dividing line and the resulting response generated by the LLM is shown below it.

Colloquial-style prompts resulted in steps that were often not relevant to the agent's situation. This strategy was also highly sensitive to the use of articles; ``the can'' in a prompt rather than ``a can'' could result in very different responses. Further, there was no readily discernible pattern in the resulting variations. 
The predicate format often results in responses with incorrect predicate arguments and generally more frequent errors. 
\cite{olmo_gpt3--plan_2021_local} also found that incorrect variable instantiation for predicates was a significant limitation in their exploration of using LLMs for agent planning. Based on these preliminary results, subsequent experiments throughout the remainder of the paper use the terse language style.

\begin{table*}
    \centering
    \begin{tabular}{p{0.12\textwidth} p{0.83\textwidth}}
    \hline 
    \emph{Prompt}     &   (EXAMPLES) (TASK) Goal: Deliver object. Task context: I am in mailroom. Aware of package addressed to Gary, package is in mailroom. Steps: \par
1. Pick up package addressed to Gary \par
2. Go to Gary's office \par
3. Put package onto desk in Gary's office (END TASK)\par 
\par(END EXAMPLES) \par
(TASK) Goal: tidy conference room. Task context: I am in conference room. Aware of can, can contents is empty, can material is metal, can property is soda, can is in conference room. Steps: 1.
\\[0.25ex] \cline{2-2} 
\noalign{\vskip 3pt}    
\emph{GPT-3} \par \emph{Response}     &  Pick up can \par
2. Take can to kitchen \par
3. Put can in recycling bin (END TASK) \\
\hline
\noalign{\vskip 3pt}    
\emph{Response} (no delimiters)     &  Take can to kitchen \par
2. Throw away can \par
3. Wash hands \\
\hline
\emph{GPT-3} \par \emph{Response} (no prompt examples)  & Remove all items from conference room. \par
2. Vacuum and sweep conference room. \par
3. Dust conference room. \par
4. Wipe down all surfaces in conference room. \par
5. Place can in recycling bin. \\
    \end{tabular}
    \caption{Example prompt with delimiters, the resulting response from GPT-3 , response when removing the delimiters from the prompt, and response when removing the example from the prompt. (All prompts: Temperature=0, model=text-davinci-001)}
    \label{tab:gpt_Example}
\end{table*}

\subsection{Delimiters}

Other researchers have shown that providing explicit tags and/or delimiters within the text of a prompt can improve the reasonableness of generated results. This outcome can be achieved without specialized fine-tuning for the task or domain \citep{reif_recipe_2022_local}, which makes it a potentially useful tool for online agent task learning. Delimiting and labeling the components of prompts encourages the LLM to generate responses more consistent with the desired structure of response, and makes it easier for the agent to extract knowledge from the LLM response.

In the template in Table~\ref{tab:prompt_template}, we introduce two kinds of delimiters. Parenthetic, paired delimiters such as ``(EXAMPLES)'' and ``(TASK)'' mark the start and end of a component. Within task descriptions, we introduce a tag delimiter to name components of the task. Delimiters are included in the GPT-3 prompt example shown in Table \ref{tab:gpt_Example}. 

The second from the bottom row of the table illustrates the response to the prompt if the delimiters are removed. The response remains reasonable, showing the flexibility of GPT-3 to different formats, but is less relevant to the embodiment of the robot and its parsing capabilities (interpretability). Delimiters also help specify stop sequences, which are used to tell GPT-3 when to stop generation. For these reasons, we adopted delimiters in the template for our primary experiments. 

\subsection{Number of examples}
Prompt examples have been shown to be effective models for shaping the desired responses of an LLM \citep[e.g.,][]{brown_language_2020,reynolds_prompt_2021_local,liu_what_2021_local,bommasani_opportunities_2021_local}. An example, presented to the LLM as part of the prompt (again, just in plain-text language), influences what a language model produces. In general, the inclusion of examples biases the LLM to produce responses that are similar to the examples. 

Given the potential utility of examples, we explore the impact of including examples in the template-based approach. A prompt example for delivering a package is depicted at the start of the prompt in Table~\ref{tab:gpt_Example}. These examples are created by a developer (not constructed by the agent). However, to-date we  use only a few distinct examples and the examples are matched to the robot's assumed ``pre-programmed'' capabilities (other examples used for experiments: ``store package'' and ``fetch printout''). Long-term, a small library of pre-defined examples could be included with a task-learning robot. 

We also evaluated prompts that did not include any examples. The last row of Table~\ref{tab:gpt_Example} shows the response to the prompt without examples. All LLM settings are the same, but the response generated contains many steps that are not relevant or interpretable. Based on early explorations in which the lack of examples provided poor results while three generic examples led to significant improvements over many different task learning prompts, we decided to include at least one example in future explorations. There is not consensus in the LLM community about what number of examples is optimal in prompt engineering \citep{liu_what_2021_local}. We used between 1 and 3 examples, and analyze the effect of the number of examples on the quality of the responses.

\subsection{Context}\label{section:context}
Providing the LLM with a representation of the situational context is necessary to achieve good situational relevance. Scope and form are two challenges to achieving an effective representation. 

Scoping decisions take into account which aspects of the situation should be included in the prompt. The full situation might include all current percepts (objects, locations, relationships, etc.), internal states, and past history. Further, because the agent is learning a novel task, it cannot use existing knowledge to limit the scope of the situation. Given these constraints, we limit exploration to the inclusion of external context and defined three different options for representing the context:
\begin{itemize}
    \item No Context: No context information is provided. This condition helps establish a baseline in that any additions of context should at least improve performance over no context.
    \item Partial context: Context is limited to the object and agent locations. This is a domain- and task-neutral heuristic that allows the LLM to attend to one item in the overall context. Partial-context templates are illustrated in Table~\ref{tab:lang_styles}. The prompt example in Table~\ref{tab:gpt_Example} uses the partial context template.
    \item Full context: All percepts are included in the prompt. This approach ensures that the LLM has access to the complete perceptual state but at the potential cost of reducing the salience of objects and relationships relevant to the task. For example, with full context, the partial task prompt illustrated in Table~\ref{tab:gpt_Example} would also include ``A large table is in conference room. A recycling bin is on floor...'' as context.
\end{itemize}

The other challenge is the form in which the context is presented. The agent has the ability to describe its environment in natural language, which is similar in form to the terse language style described above. We initially explored a colloquial and predicate representation of the context, matching the language style of the prompt used; however, experimentation proved the benefits of the terse format, so we settled on a form that matches the terse language format of the prompt.

\subsection{Object properties}\label{section:object-features}
Within the context, objects have associated properties, such as the material of a cup is paper. Which object properties should be included in the context? We explored two conditions: full (all perceptual features, the \texttt{feature-list} as in Table~\ref{tab:lang_styles}, of the target object are included in the prompt) and partial/none (only the object name is included). The feature list for each object was presented in Table~\ref{tab:conference_room_objects}. Again, as for context, in the ideal case the agent would include only properties relevant to the task, but, a priori, it cannot know which properties are relevant.

Because some properties strongly influence how an object relates to the task (a glass cup vs. a paper cup), we generally expect that a full set of properties would improve reasonableness and relevance, but long property lists may have a diluting effect on the LLM's ability to pick out salient properties relevant to the task.

\subsection{Temperature}

We also analyzed the affect of temperature on the measures. Temperature ranges from 0 to 1. At temperature=0, GPT-3 produces (largely) deterministic results: it always produces the token with the highest probability as it generates responses. At higher temperatures, there is greater variance in the text that is generated. In general, as temperature increases, the LLM chooses tokens the next token in the sequence less predictably. We generally expect that temperature=0 responses will have higher reasonableness than higher temperature responses to the same prompt. Because a specific situation might require a less common response, however, we used other temperatures (0.3, 0.8) to explore how different temperatures influence relevance.

\section{Experimental Methodology and Results}
In this section we describe the experiment methodology and results comparing the prompt construction strategies and variations as outlined in the previous section. Further, because much of the initial explorations were done using the ``tidy conference room'' task, we also introduce two other tasks to mitigate the likelihood that the chosen strategies are specific to the task/context combination. The first, ``tidy kitchen,'' is a similar task but in a different context (kitchen vs. conference room). The other task is ``prepare conference room for banquet.'' This task is performed in the same context (conference room), but is a different task. This task also has the advantage of being uncommon and thus more representative of how an LLM could support learning idiosyncratic tasks.

The goal for the experiments is to learn what effect variations in the prompt have with regards to the interpretability, relevance, and reasonableness of responses. The resulting measures will then inform if and how an LLM could be used by an agent for task learning. Ideally we would test fully automatically over a large dataset of tasks, domains, objects, features, prompt examples, and LLM settings. However, because we are searching for ``acceptable'' variations that an agent might be able to use, we cannot simply automate the search (given available resources for evaluation). Rather than a fully automated approach, each response was evaluated manually along the three dimensions introduced above. Three raters (all authors of this paper) evaluated each LLM response (roughly 400 in all).\footnote{When there was disagreement among the raters, those items were discussed until consensus was reached.} 

Evaluation rubrics were developed for each measure. A response was rated ``reasonable,'' if the raters could imagine the response represented a valid set of steps in some possible task environment. For example, for the ``metal can'' in the conference room, reasonable responses included ones in which the can was put in the trash or put in the recycling bin. For situational relevance, we defined a ``gold standard'' for each object that represented the desired action in the specific context of the task. Because there is a ``recycling bin'' present in the conference room, steps generated by the LLM that result in the can being placed in the trash were not rated as relevant to the agent's specific situation. Finally, the interpretability of individual responses drew on the author's knowledge of the agent's native parsing and grounding capabilities. A simple, concrete response such as ``pick up can,'' is readily interpreted by the agent. However, vocabulary unknown to the agent, abstract concepts that cannot be readily grounded (e.g., ``appropriate'' in ``Put objects in appropriate places''), and complex syntactic constructions were all deemed ``not interpretable.'' 

Due to the time required to perform evaluations and knowledge required to make accurate judgements, manual evaluation was a limiting factor in our ability to perform larger-scale experiments that would explore more variations as well as larger sets of objects, contexts, and tasks. Future experiments could possibly rely on a crowd-sourcing service for evaluation now that we have established clear guidelines for evaluation. However, these smaller-scale experiments provide sufficient data to analyze effects of prompt variations for a small set of task domains, informing areas for future exploration.
    
A set of objects (and features) were defined for each task/domain pair (objects for ``tidy conference room'' were presented in Table \ref{tab:conference_room_objects}). The other task domains include similar common objects, such as a carton of milk and a box of napkins, and some objects are used in multiple task contexts (a ``plate'' in both the kitchen and banquet tasks; a ``paper cup'' in both the kitchen and conference room, etc.).  In each task domain, we varied the number of prompt examples (1-3) and the LLM temperature (0.0, 0.3, 0.8). For temperatures greater than 0, we retrieved the best 3 responses from the LLM. All experiments were conducted using GPT-3 model \emph{text-davinci-001}.

The primary results were generated using the terse language style, full object features, and partial situation context. A secondary experiment that addresses variations on the provided context and object features is discussed in section \ref{section:result-context}.

\begin{figure}[t]
\vskip 0.05in
\begin{center}
\begin{minipage}{.28\textwidth}
        \centering
        \includegraphics[width=0.99\linewidth]{"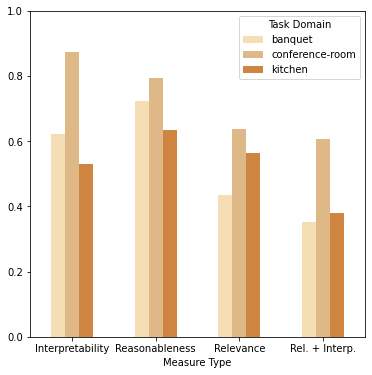"}
        \caption{Across Domains}
        \label{fig:results_domains}
    \end{minipage}%
    \hspace{.05\textwidth}
\begin{minipage}{.28\textwidth}
        \centering
        \includegraphics[width=0.99\linewidth]{"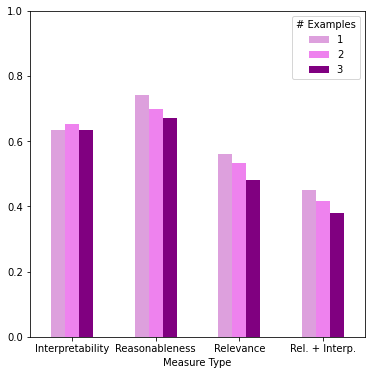"}
        \caption{Num. examples}
        \label{fig:results_examples}
    \end{minipage}
    \hspace{.05\textwidth}
    \begin{minipage}{.28\textwidth}
        \centering
        \includegraphics[width=0.99\linewidth]{"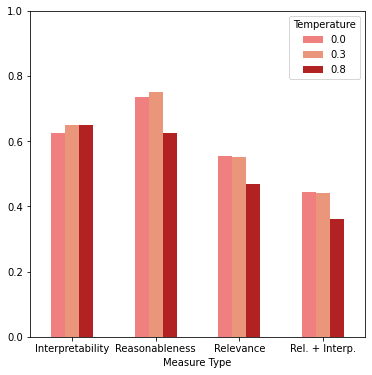"}
        \caption{Temperature}
        \label{fig:results_temperature}
        \end{minipage}
            
\end{center}
\vskip -0.2in
\end{figure}

Figure~\ref{fig:results_domains} shows the overall results for each task domain for all variations tested. For all results we report the mean percentage of responses that are reasonable, interpretable, and situationally relevant. This figure, and others, also includes the percentage of responses that were both situationally relevant and interpretable. This condition reflects the percentage of responses that the agent could use to learn the task successfully with no other support.  Over 60\% of the responses for the ``tidy conference room'' domain were both situationally relevant and interpretable, compared to around 40\% for the two other task domains. These results show significant opportunity to exploit LLMs for agent task learning, but also need for further improvement. Subsequent sections highlight how variations in the prompting strategy can improve the interpretability, relevance, and reasonableness of responses. 

\subsection{Number of Examples}
Figure \ref{fig:results_examples} summarizes the impact of the number of prompt examples across all three task domains on the evaluation dimensions. These results show, with the exception of interpretability, which remained relatively stable, that increasing the number of prompt examples has a negative effect on the quality of the responses generated. One example appears sufficient to influence the LLM to generate responses with the desired format and content, while additional examples mainly serve as distractors to the current task. However it could also be the case that certain prompt examples were better for some tasks or task objects. To further investigate, we plan to perform larger experiments where we test a wider range of prompt examples with a larger set of objects in more task domains.

\subsection{Temperature}
Figure~\ref{fig:results_temperature} shows the impact of temperature across all task domains on the evaluation dimensions. The percentage of interpretable responses remained relatively consistent across different temperatures. Temperatures of 0~(near-deterministic) and 0.3 provided comparable results, but temp=0.8 has a negative effect on reasonableness and situational relevance. 
    
At higher temperatures, more varied responses are generated by the LLM, which can be beneficial for an agent that may want to evaluate many different alternatives, but variation (unsurprisingly) increases the probability that some of those responses are not relevant or reasonable. Thus, there is a trade-off between number of unique responses from the LLM and the relevance/reasonableness of the response, that will need to be explored by a cognitive agent using LLM responses to learn tasks.

\begin{figure}[t]
\vskip 0.05in
\begin{center}

\begin{minipage}{.44\textwidth}
        \centering
        \includegraphics[width=0.7\linewidth]{"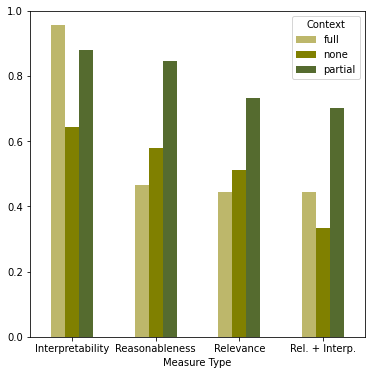"}
        \caption{\label{hdfs} \protect\raggedright Results from small experiment showing impact of context variations}
        \label{fig:results_context}
    \end{minipage}%
    \hspace{.05\textwidth}
    \begin{minipage}{.44\textwidth}
        \centering
        \includegraphics[width=0.7\linewidth]{"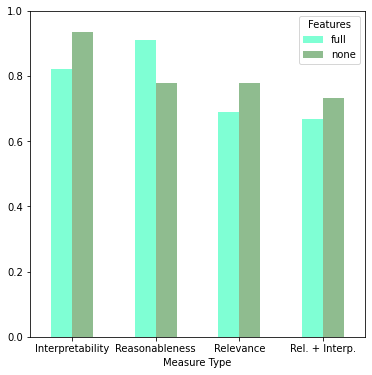"}
        \caption{\label{hdfs} \protect\raggedright Results from small experiment showing impact of feature variations}
        \label{fig:results_features}
    \end{minipage}%
\end{center}
\vskip -0.2in
\end{figure}

\subsection{Context}\label{section:result-context}
Through exploration, we initially identified that including too much or too little context negatively impacted the results. Therefore in our primary experiment across the three task domains, we only explored the variation of partial context, that includes the agent's location and the object location. Similarly we only explored the variation of providing object features as context in the prompt. This variation with partial context and object features can be observed in Table \ref{tab:gpt_Example}.

However we conducted a small additional experiment to directly examine the effect of variations in context and features provided. The experiment parameters are the same as the larger experiment, but limited to the tidy conference room task, using only one prompt example. Figure \ref{fig:results_context} shows the effect of context along the evaluation criteria. Definitions of the three conditions, none, partial, and full context, are given in section \ref{section:context}. With the exception of interpretability, the partial context variation far outperforms the others. Our hypothesis is that the full context provides too many distracting objects that are not relevant to each individual task, and no context misses key details, such as that a mug is in the sink (instead of on the table). This is especially the case with our task domains, where achieving the task does not involve many interactions between objects. We suspect with tasks that have more object interactions, such as assembling furniture, it will perform better with a fuller context of the situation.

Figure \ref{fig:results_features} shows the impact of including object features for the variations of none and full features (defined in section \ref{section:object-features}). Surprisingly the no feature variation slightly outperformed full features on all dimensions, except for reasonableness. We hypothesize that this is because many of the features are not relevant to the specific task at hand, such as that a mug is ceramic. The result of including full features is that more responses are reasonable but they are not the desired response that is situationally relevant. We plan to investigate this issue further by presenting features in different formats in the prompt, and performing larger scale experiments across more task domains, objects, and features.

\begin{wrapfigure}{r}[-5pt]{0.35\textwidth}
        \vspace*{-\topsep}
        \centering
        \includegraphics[width=2in]{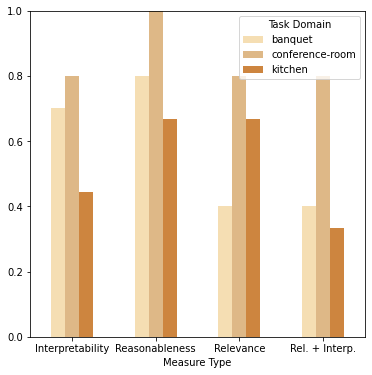}
        \caption{\label{hdfs} \protect\raggedright Results of best variation in primary experiment (Temp=0; 1 prompt example).}
        \label{fig:results_best_variation}
\end{wrapfigure}

\subsection{Discussion}
Our experiment provides evidence that appropriate prompt examples can bias an LLM to produce interpretable, relevant responses and only a single example is needed. Increasing temperature increases the number of unique responses generated that the agent can consider, but has a slight negative impact on relevance and reasonableness. Context, including object features, is beneficial, but insufficient, for producing the most relevant steps. Context can be distracting, sometimes reducing situational relevance.

The best results from the primary experiment were achieved with 1 prompt example and temp=0. Results for only this variation are shown in Figure \ref{fig:results_best_variation}. For ``tidy conference room,'' over 80\% of responses were both relevant and interpretable. The other tasks show some promise for using LLMs as a knowledge source for task learning, with 60-80\% reasonableness for a wholly novel task.

\section{Additional Prompt Engineering Methods}

Below we outline next steps based on the results of the experiment. These efforts seek to further involve the agent in the prompting and generation process.  While further prompt engineering can improve performance, we are also exploring integrating multiple sources of knowledge, including human interaction and planning, so that the agent need not rely on just the LLM \citep{AAAIpaper_arxiv_2022}.

\subsection{Interactive Tuning of Generation}\label{section:iterative}

The prompting strategy discussed above generated a prompt and then received a response from the LLM. However, the LLM generates responses one token (``word'') at a time. We investigated a more interactive approach to response generation, allowing the agent to bias the generation towards (action) words that it knows. Rather than making a single request for a response, a single word is requested (using temperature 0) following each step number. The first word after a step number is (almost) always an action verb using our prompt template. Using the top\char`_logprobs feature of the GPT-3 API, the top N responses and their relative probability can be retrieved.

For example the top five words generated after ``1.'' from Table \ref{tab:gpt_Example} are ``Pick''(48\%), ``Take''(40\%), ``Remove''(3\%), ``Throw''(1.6\%), and ``Put'' (1.4\%). Using a list of the words that the agent knows (e.g., ``Pick'') it chooses known words above a minimum threshold of 10\% and then resends the prompt with that added word (``1. Pick''). When no known words exceed the threshold, other words are chosen that are above a higher threshold of 60\%. This process results in multiple responses to consider for each step. The process is repeated for each step for the first word after the step number until the LLM generates ``(END TASK).''

\begin{figure}[t]
\vskip 0.05in
\begin{center}

\begin{minipage}{.44\textwidth}
        \centering
        \includegraphics[width=0.7\linewidth]{"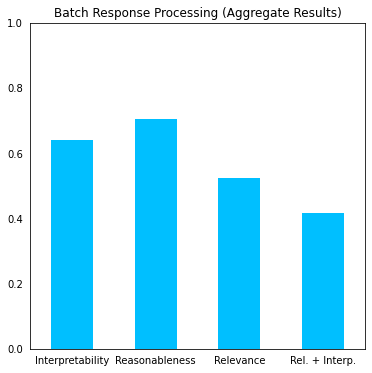"}
        \caption{Aggregate results (all domains, all variations) for the original (or ``batch'') processing of responses from GPT-3.}
        \label{fig:results_aggregate_batch}
    \end{minipage}%
    \hspace{.1\textwidth}
    \begin{minipage}{.44\textwidth}
        \centering
        \includegraphics[width=0.7\linewidth]{"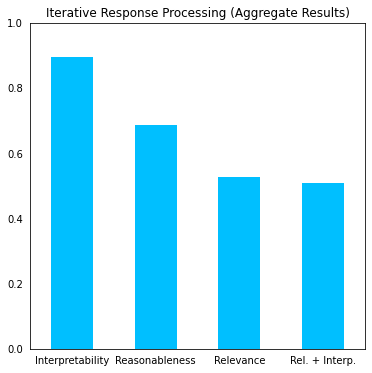"}
        \caption{Aggregate results (all domains, all variations) for the revised (or ``iterative'') processing of responses from GPT-3.}
        \label{fig:results_aggregate_iterative}
    \end{minipage}%
\end{center}
\vskip -0.2in
\end{figure}

We repeated the primary experiment, using the same parameters, over the three task domains using this new iterative tuning strategy. Aggregate results for the primary experiment and this new iterative experiment are shown in Figures \ref{fig:results_aggregate_batch} and \ref{fig:results_aggregate_iterative}. Unsurprisingly, the iterative strategy leads to greater interpretability, and does not negatively influence the other measures. While a modest step, these improved results suggest that more interactive prompt generation has potential to improve all the measures (e.g.,  biasing responses towards objects and features currently observed to attempt to improve situational relevance).

\subsection{Complementary Representations of Tasks}\label{section:goals}

The prompt template we used in the primary experiment focuses on generating actions. However, agents often use both goal- and action-representations to represent tasks \citep{mininger_expanding_2021}. A goal representation enables an agent to use planning/search capabilities to discover and/or verify the steps to achieve the goal. Our own work shows how a goal representation reduces reliance on receiving complete and correct steps from the LLM \citep{AAAIpaper_arxiv_2022}. 

Prompting the LLM to generate the goal also provides context for each generated step, a process referred to as ``chain of thought reasoning'' in the LLM community \citep{wei_chain_2022_local}. To elicit task goal knowledge from the LLM, we modified the prompt template and prompt examples to include a goal description delimited with ``(RESULT)'' and ``(END RESULT).''
Using this strategy for a plastic bottle in a conference room, GPT-3 generates ``The goal is that the plastic bottle is in the recycling bin,'' which is both relevant and interpretable. Informal findings to-date suggest the LLM usually generates goals that are interpretable and are often relevant. We plan to explore eliciting goal knowledge formally to evaluate its effect on overall relevance and interpretability.

\section{Conclusions}

LLMs offer great potential for online agent knowledge acquisition. However, LLM responses are sensitive to the content and form of each prompt. Thus to use the LLM effectively for online task learning, an agent must construct prompts that engender the LLM to produce responses that it can interpret with its native language facility and that are relevant to its current situation. The prompt construction strategies we report in this paper suggest that an agent can achieve this goal via a template-based approach agnostic to task and domain.

As suggested previously, future work is proceeding in two complementary directions. In the first, we are investigating the use of multiple knowledge sources, including human feedback and planning, to learn tasks quickly and robustly. This work mitigates reliance on the LLM as the only source of knowledge and may lower the criteria for acceptable/good performance from the LLM. In the second thrust, based on the success of the preliminary work on more interactive prompt generation, we will continue to explore further improvements to the prompting strategy, such as biasing generation toward vocabulary the agent understands and objects it can observe.

\begin{acknowledgements} 
\noindent
This work was supported by the Office of Naval Research, contract N00014-21-1-2369. The views and conclusions contained in this document are those of the authors and should not be interpreted as representing the official policies, either expressed or implied, of the Department of Defense or Office of Naval Research. The U.S. Government is authorized to reproduce and distribute reprints for Government purposes notwithstanding any copyright notation hereon. We acknowledge the feedback of the anonymous reviewers who effectively highlighted sections where additional clarity was needed.
\end{acknowledgements} 

{\parindent -10pt\leftskip 10pt\noindent
\bibliographystyle{cogsysapa}
\bibliography{transoar,local}

}

\end{document}